%% file: paper.tex
\documentclass[letterpaper, 10 pt, conference]{ieeeconf}
\IEEEoverridecommandlockouts    
\input{stachnisslab-latex}

\input{stachnisslab-math}

\usepackage{graphicx}  
\usepackage{float}  
\usepackage{subfigure}  
\usepackage{cite}
\usepackage{multirow}
\usepackage{url}

\title{\LARGE \bf Hybrid Map-Based Path Planning for Robot Navigation \\ in Unstructured Environments}

\author{Jiayang Liu \and Xieyuanli Chen  \and Junhao Xiao$^*$ \and  Sichao Lin \and  Zhiqiang Zheng \and  Huimin Lu$^*$ 
  \thanks{All authors are with the College of Intelligence Science and Technology, National University of Defense Technology, Changsha, China.}%
  \thanks{$^*$corresponding authors, junhao.xiao@ieee.org, lhmnew@nudt.edu.cn}
  \thanks{This work was supported in part by the National Science Foundation of China under Grant U1913202, and U22A200600, as well as Major Project of Natural Science Foundation of Hunan Province under Grant 2021JC0004.
  }%
}

\begin{document}
\maketitle
\thispagestyle{empty}
\pagestyle{empty}

\begin{abstract}
Fast and accurate path planning is important for ground robots to achieve safe and efficient autonomous navigation in unstructured outdoor environments. However, most existing methods exploiting either 2D or 2.5D maps struggle to balance the efficiency and safety for ground robots navigating in such challenging scenarios. In this paper, we propose a novel hybrid map representation by fusing a 2D grid and a 2.5D digital elevation map. Based on it, a novel path planning method is proposed, which considers the robot poses during traversability estimation. By doing so, our method explicitly takes safety as a planning constraint enabling robots to navigate unstructured environments smoothly.
The proposed approach has been evaluated on both simulated datasets and a real robot platform. The experimental results demonstrate the efficiency and effectiveness of the proposed method. Compared to state-of-the-art baseline methods, the proposed approach consistently generates safer and easier paths for the robot in different unstructured outdoor environments. The implementation of our method is publicly available at \url{https://github.com/nubot-nudt/T-Hybrid-planner}.

\end{abstract}

\section{Introduction}
\label{sec:intro}

Safe and efficient autonomous navigation is essential for ground robots operating in challenging unstructured outdoor environments. Path planning is a key component of an autonomous navigation system. 
In unstructured environments, robots have to deal with not only common obstacles, such as trees and rocks, but also the challenges of special terrains, such as steep slopes and hollows. 
In such cases, path planning should consider both terrain constraints and kinematics constraints of the robot.

Several studies based on 2D maps have been proposed for outdoor path planning~\cite{stentz1994icra,Elfes1989UsingOG,wen2021icra}. 
In those approaches, each perceived grid can only be classified as either occupied or free; in other words, such binary classification lacks of evaluating the underlay traversability cost. 
As a result, such 2D map-based methods are weak in describing terrain, therefore may introduce difficulties for robot path following. 
On the other hand, digital elevation maps (DEMs)~\cite{fankhauser2014robot} generally have terrain information such as height, roughness, and normal vectors. 
Therefore, the generated paths in 2.5D DEMs~\cite{Wermelinger2016iros} are safer and easier to pass than the 2D competitors. 
However, as DEMs cannot reasonably represent overhanging structures such as trees and bridges, it is time-consuming to calculate the traversability of 2.5D terrain during online navigation. 

\begin{figure}[t]
	\centering
	\vspace{0.25cm}
	\includegraphics[width=\linewidth]{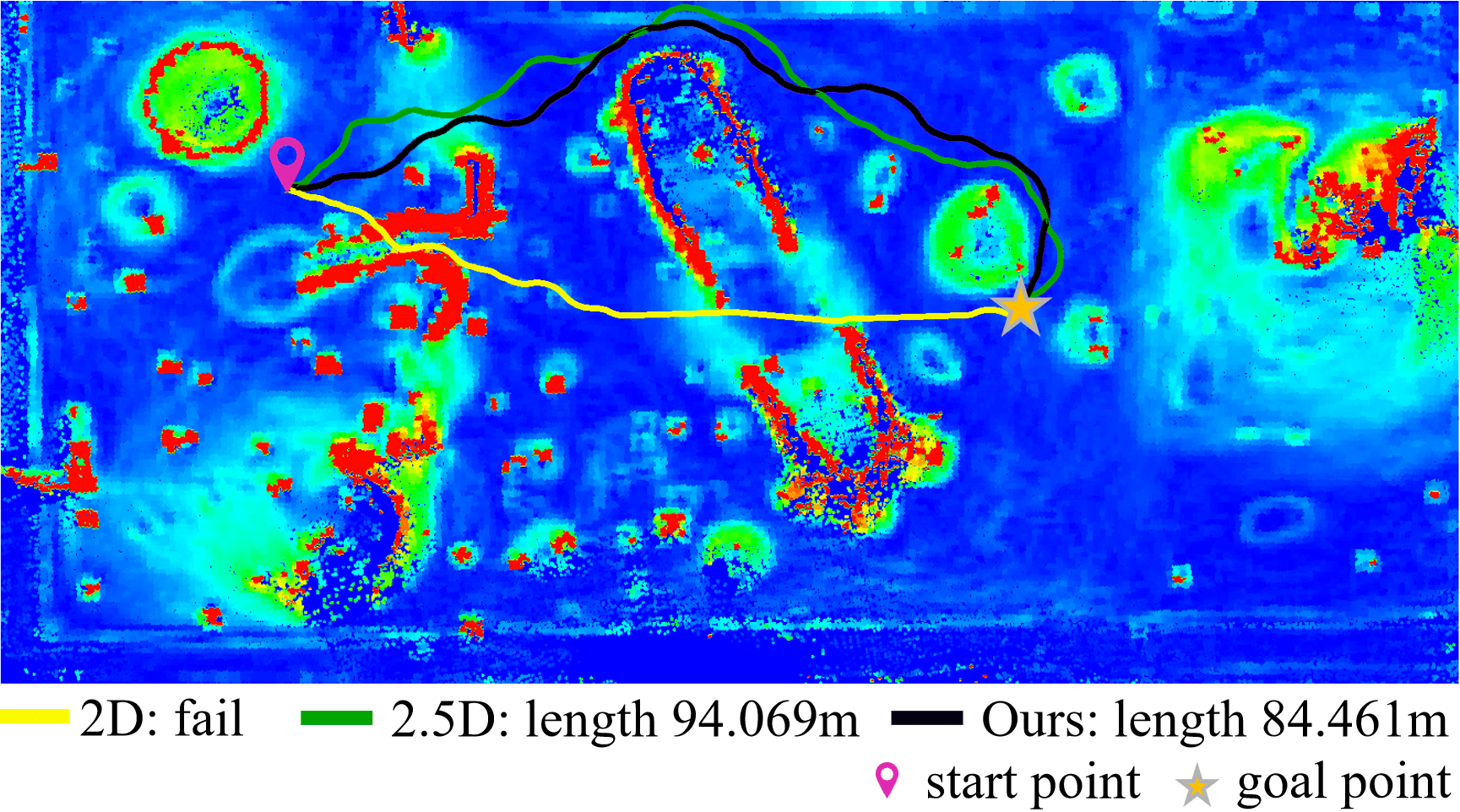}
	\caption{Comparison of path planning in a Mars-like surface dataset. The color map is a hybrid map generated by our terrain assessment. The color represents traversability, which gradually increases from warm to cool colors. The red-colored area represents impassable 2D map areas, and the rest belongs to the 2.5D map. The black path is generated by our method, while the yellow and green paths are from a 2D map-based and a 2.5D map-based baseline methods, respectively. The corresponding path lengths are listed below. As can be seen, our method generates a shorter path across safer areas.}
	\label{fig:first}
\end{figure}

To tackle the problems mentioned above, in this paper, we propose a novel path-planning method in unstructured outdoor environments. 
More specifically, a sing-layer dynamic voxel grid is employed to segment the raw point clouds, where the absolutely impassable areas are marked as occupied. 
The remaining areas are further modelled in the 2.5D layer. 
Afterwards, the traversability of each DEM grid is calucated according to its roughness and the robot's possible orientation.
By considering the cost of terrain traversability in path planning, a shorter and safer path can be eventually generated, as shown in~\figref{fig:first}.
The proposed method has been evaluated based on a publicly available dataset and our real ground robot platform. 
The experimental results show that our method outperforms state-of-the-art planning methods in unstructured outdoor environments.

In sum, the contributions of our methods are threefold:
(i) Our method assesses various complex terrain accurately and creates 2D+2.5D hybrid map representations possessing the advantages of 2D and 2.5D maps;
(ii) A novel path planning method T-Hybrid A* has been proposed, which generates better and safer paths compared to existing state-of-the-art methods;
(iii) A novel navigation method has been proposed for ground robots based on the proposed hybrid map and T-Hybrid A*, enabling our real robot to navigate safely and efficiently in real unstructured terrains.
%

\section{Related Work}
\label{sec:related}

For ground mobile robots working in unstructured outdoor environments, representing the terrain traversability in a map is a key issue of path planning \cite{papadakis2013terrain}. 
Existing map representations for outdoor path planning mainly include four different groups: 2D grid map \cite{Elfes1989UsingOG}, 2.5D elevation maps \cite{choi2012urai}, 3D voxel grids \cite{wurm2010icra}, and point cloud maps \cite{gao2019jfr}.

Occupancy Grid Map is a frequently used 2D map representation for path planning. 
According to the occupancy probability, the detected grids are divided into occupied or free grids, while the undetected grids are represented as unknown grids~\cite{elfes1990uai}.
In the unstructured outdoor environment, obstacle areas such as bumps, pits, ramps, and gullies are considered, and the occupied grid map is generated by geometric or numerical calculation for path planning~\cite{Guan2021ral}.
Demonstration learning has been employed to calculate terrain costs in complex unstructured terrains in \cite{silver2010ijrr}, where D*  \cite{stentz1994icra} is used for path planning. 
Overby~\etalcite{overbye2020icra} exploited a buffer combined obstacle detection with a terrain gradient map to generate a 2D grid map, then used a multi-step path planning method to realize autonomous navigation in unknown and off-road environments. 
Hybrid A* is used for path planning on a 2D occupancy grid map in \cite{guan2021rss}. 
Hine~\etalcite{hines2021ral} constructed virtual grids and assigned certain traversability values to virtual areas. Then they projected the 3D grids to the 2D plane, and planned a path using Hybrid A*. 
The construction of 2D grid maps is fast, and the terrain traversability can be queried quickly during path planning. 
However, their simplicity also makes it hard to evaluate whether a robot will be dangerous in unstructured outdoor environments, such as small bumps and slopes (the robot can cross such non-obstacle areas in a unique orientation), thus unreliable for a robot navigating in unstructured outdoor terrains.

As 3D maps, both raw point clouds and 3D voxel grids can describe the terrain more comprehensively than 2D maps. 
The point clouds are the raw sensor data format of a Lidar sensor, which can describe the environment information relatively densely based on robotic mapping techniques.
Reina~\etalcite{reina20143sr} proposed an uneven point descriptor to analyze the local point cloud normal vector, calculated the traversability index of the region, and distinguished the traversable region from the impenetrable region. 
Kr{\"u}si~\etalcite{krusi2017jfr} carried out path planning in an unstructured outdoor environment based on a sampling method directly on the raw point clouds. 
Liu~\cite{liu2016tcyb} used a tensor voting framework to plan a safe path on point cloud online. 
However, incorporating terrain assessment into path planning will increase the planning time significantly in large-scale outdoor environments due to the huge amount of point cloud data. 
Octomap~\cite{hornung2013Oar} is a common 3D voxel grid representation that subdivides the 3D space selectively according to the probability distribution of obstacles.
It is very important for path planning of aerial robots~\cite{dang2019icar}, robot arm motion planning~\cite{wang2019ral}, and 3D path planning of ground robots in multi-layer environments. 
Hertl~\etalcite{hertle2013ecmr} extracted drivable surfaces on 3D terrain octmap and used motion primitives on surface cells to realize path planning for the tracked robot in a 3D complex environment.
A traversability map is generated based on octomap, which is then employed for safe and efficient path planning \cite{yu2019rcar}. 
However, similar to 3D point cloud maps, octomap has large storage space for terrain information, so it is slow to query terrain information during path planning. 
Besides, the ground robot does not need to navigate through multi-layer structures in an unstructured outdoor environment. 
Therefore, using 3D maps for path planning in unstructured outdoor terrain is inefficient.

The digital elevation map is a 2.5D form of terrain representation, which realizes the digital simulation of terrain through limited elevation data. 
Compared with a 3D grid map, an elevation map is equivalent to the compression of terrain, which is more conducive to efficient path planning in the outdoor environment. 
2.5D digital elevation map usually contains multi-level terrain information, such as 2D position, slope, roughness, normal vector, and other information. 
It has been well applied in the path planning of complex outdoor terrain. 
Sing~\etalcite{singh2000icra} rasterized the terrain data collected by the stereo camera, calculated the orientation angle and roughness to generate an accessible grid map, and used D* global planner and arc sampling local planner to navigate on the Martian terrain. 
Dail~\etalcite{daily1988icra} used elevation maps to plan the global guidance route in the off-road environment and provided subpoints for real-time reactive obstacle avoidance planning.
Fan~\etalcite{fan2021rss} established a 2.5D traversable risk map by considering the uncertainty factors. They used an efficient MPC framework to solve the optimal control problem of nonconvex constraints in complex terrain.
Fankhauser~\etalcite{fankhauser2016universal} opened source a tool kit for calculating elevation maps. They estimated traversability by calculating each terrain grid's height, roughness, slope, and other information, then used the RRT algorithm to plan a safe path for legged robots.  
However, the construction of elevation maps is complicated. Robots do not need elevation information in complex unstructured outdoor terrain with steep slopes, gullies, stone piers, and other impassable areas, so the navigation in 2D map is faster and more convenient.

To fuse the advantages of 2D maps and 2.5D maps, this paper proposed a novel hybrid map representation. 
Firstly, the absolutely impassable areas are only represented in 2D, where the rest are further processed to build a 2.5D map. 
During path planning, impassable regions are not considered.
In other words, the robot's safety is only evaluated in the 2.5D map. Therefore, the computation cost of path planning is reduced. 
At the same time, a T-Hybrid A* algorithm is proposed for path planning on the hybrid map considering terrain constraints.

\section{Our Approach}
\label{sec:main}

\begin{figure}[t]
  \centering
  \includegraphics[width=1.0 \linewidth]{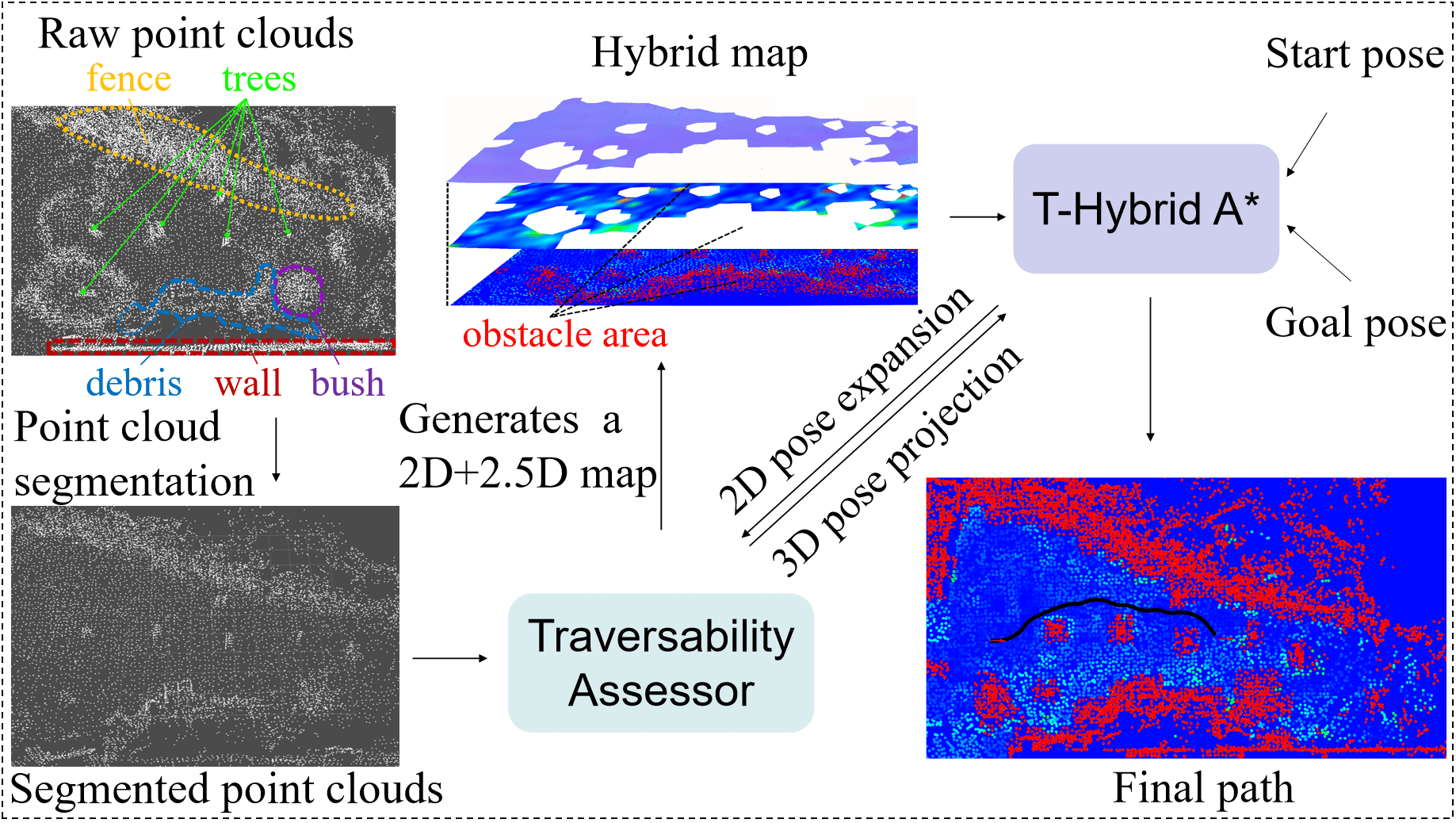}
  \caption{Pipeline overview of our path planning method. It first uses the dynamic voxel grids to segment the raw point clouds, then the traversability assessor calculates terrain traversability from the segmented point clouds and generates a hybrid map. In path planning, our terrain assessment method constantly calculates the 3D orientation of the robot projected on the terrain constantly. In the end, taking the start pose, goal pose, and hybrid map as inputs, the final safe path is generated by our T-Hybrid A* algorithm.}
  \label{fig:system}
\end{figure}

\figref{fig:system} shows the architecture of the proposed path-planning approach. 
As shown in the figure, the point cloud is segmented with dynamic voxel grids, whereafter the overhanging structure such as trees are filtered (see~\secref{sec:pcs}). 
Afterwards, a two-phase terrain assessment is carried out.
In the first phase, each voxel grid is fitted with a local plane, which is then employed to calculate the roughness and traversability of the terrain. As a result, a hybrid map can be built (see~\secref{sec:ta}). 
In the second phase, the robot's 3D orientation is calculated in $SO(3)$. 
In the end, the real terrain traversability is calculated according to the roughness, pitch angle, and roll angle of the robot, which are exploited by our proposed T-Hybrid A* to generate safe and efficient paths (see~\secref{sec:tha}). 

\subsection{Point Cloud Segmentation} 
\label{sec:pcs}
\begin{figure}[t]
  \centering
  \includegraphics[width=0.6 \linewidth]{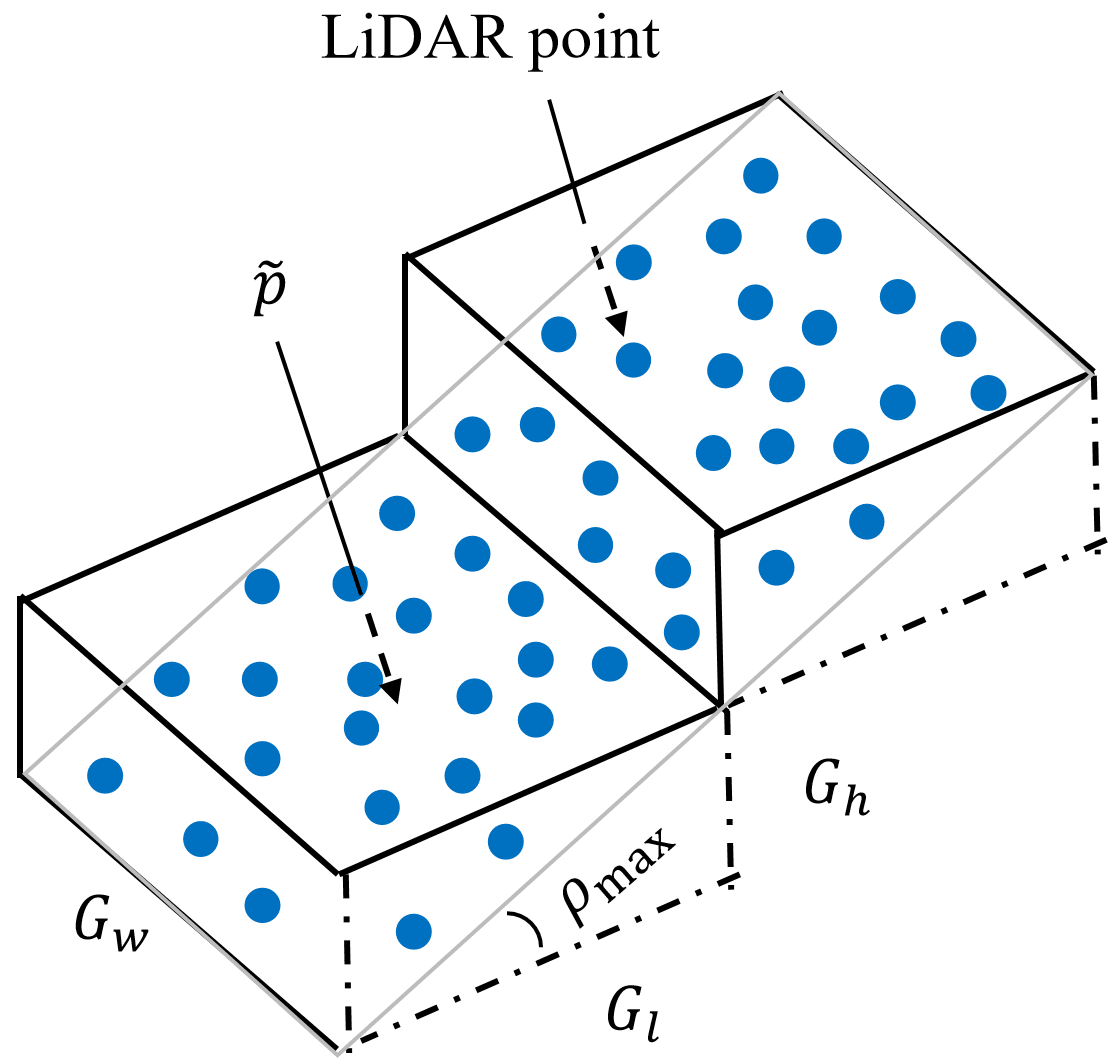}
  \caption{Point cloud segmentation based on dynamic voxel grids. The height of the center point $\widetilde{p}$ of each voxel grid changes with the height of the ground surface.}
  \label{fig:grid}
\end{figure}

Due to a large amount of point cloud data, the speed of terrain assessment is greatly affected. 
The general practice is to use a voxel grid filter to reduce the point cloud density. 
However, it cannot segment the overhanging structures such as trees and bridges reasonably~\cite{jiayang2022ccc}. 
We therefore employ the dynamic voxel grid that extracts only the ground surface points affecting the robot's movement. In this case, the overhanging structures are ignored during terrain assessment, where the computation cost is reduced. 
As shown in~\figref{fig:grid}, the blue point cloud represents a small slope. 
In point cloud segmentation, the size of the black voxel grid needs to cover the whole robot, which can be calculated by:
\begin{align}
G_l = G_w = \sqrt{R_l^{2}+R_w^{2}}\label{eq:1}, \\
G_h = G_l\tan \rho_{max} \label{eq:2},
\end{align}
where $G_l$, $G_w$, and $G_h$ are the length, width, and height of the voxel grid.
$R_l$ and $R_w$ are the length and width of the robot, and $\rho_{max}$ is the maximum slope that the robot can pass through. 
As the area with a slope higher than $\rho_{max}$ is impassable for the robot, the height of the voxel grid is set to $G_h$.  
As a result, the overhanging structures are excluded from the voxel grids.

The height of the geometric center $\widetilde{\v{p}}$ of each 
voxel grid changes with the ground surface. 
Therefore, the dynamic voxel grids can model the terrain with only single-layer grids, thus removing the effect of overhanging structures on terrain traversability analysis. 
At the same time, less point cloud data can improve the efficiency of terrain evaluation.

\subsection{Terrain Assessment}
\label{sec:ta}
For ground mobile robots, excessive pitch angle and roll angle may cause the robot to overturn. 
Therefore, terrain traversability analysis should consider the 3D orientation of the robot at each footprint. 
In this case, a local plane is fitted in each voxel grid, where the robot is projected on. 
After the projection, the robot orientation can be calculated, assuming the robot is attached to such a local plane. 

\begin{figure}[t]
  \centering
  \includegraphics[width=0.75\linewidth]{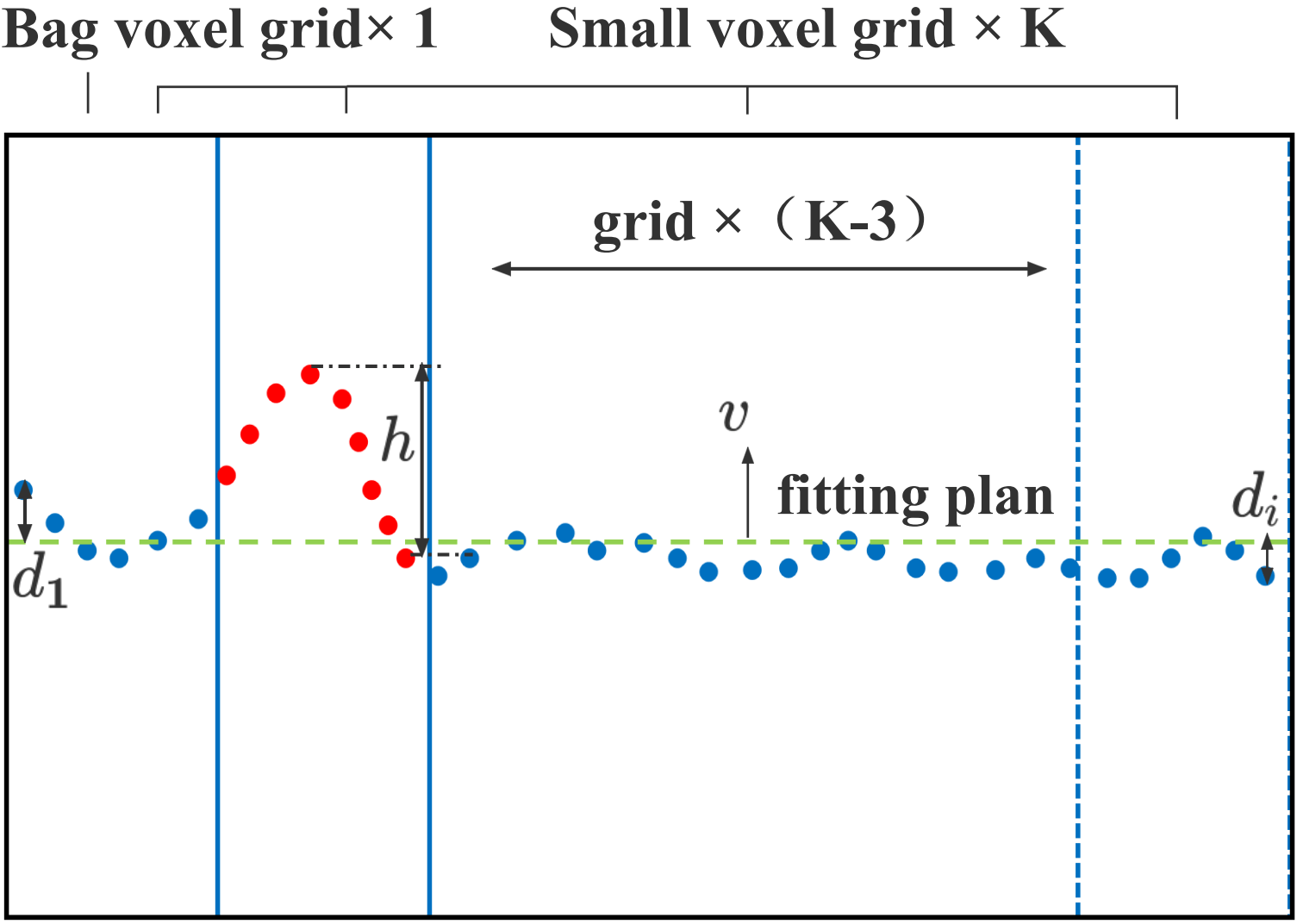}
  \caption{Schematic diagram of terrain roughness calculation. The red point clouds are identified as obstacles. The black rectangle is a large voxel grid, while the blue rectangles are small. The green dotted line in the figure represents the fitted plane.}
  \label{fig:assessment}
\end{figure}

To improve the accuracy of terrain evaluation while ensuring the evaluation speed, we use multi-resolution voxel grids. \figref{fig:assessment} is a schematic diagram for calculating terrain roughness. 
We first fit a local plane according to the points in the large black voxel grid. 
The geometry centroid $\widetilde{\v{p}}$ and covariance matrix $\m{C}$ in each voxel grid is calculated by
\begin{equation}
\widetilde{\v{p}} = \frac{1}{N}\sum_{k = 1} ^N \v{p}_k \label{3}, 
\end{equation}
\begin{equation}
\m{C} = \sum_{k = 1}^N(\v{p}_k-\widetilde{\v{p}})(\v{p}_k-\widetilde{\v{p}})^T \label{4},
\end{equation}
where $N$ is the size of point clouds in the grid. 
Principal component analysis (PCA) \cite{hoppe1992surface} is employed to calculate the minimum eigenvector as plane normal vector $\v{v}$.
The slope $\rho$ is calculated according to the acute angle between $\v{v}$ and the horizontal plane.
If $\rho > \rho_{max}$, the whole area is marked as 2D obstacle map.
This method can detect steep slopes, walls, and so on. 
Otherwise, the point clouds in the large voxel grid are subdivided by the blue small voxel grid, and the resolution of the small voxel grid is set empirically. 
If the resolution is too small, there may be no point clouds within the small voxel grid, which is inefficient. Conversely, terrain information is inaccurately represented if the resolution is too large.

\begin{figure}[t]
  \centering
  \includegraphics[width=1.0\linewidth]{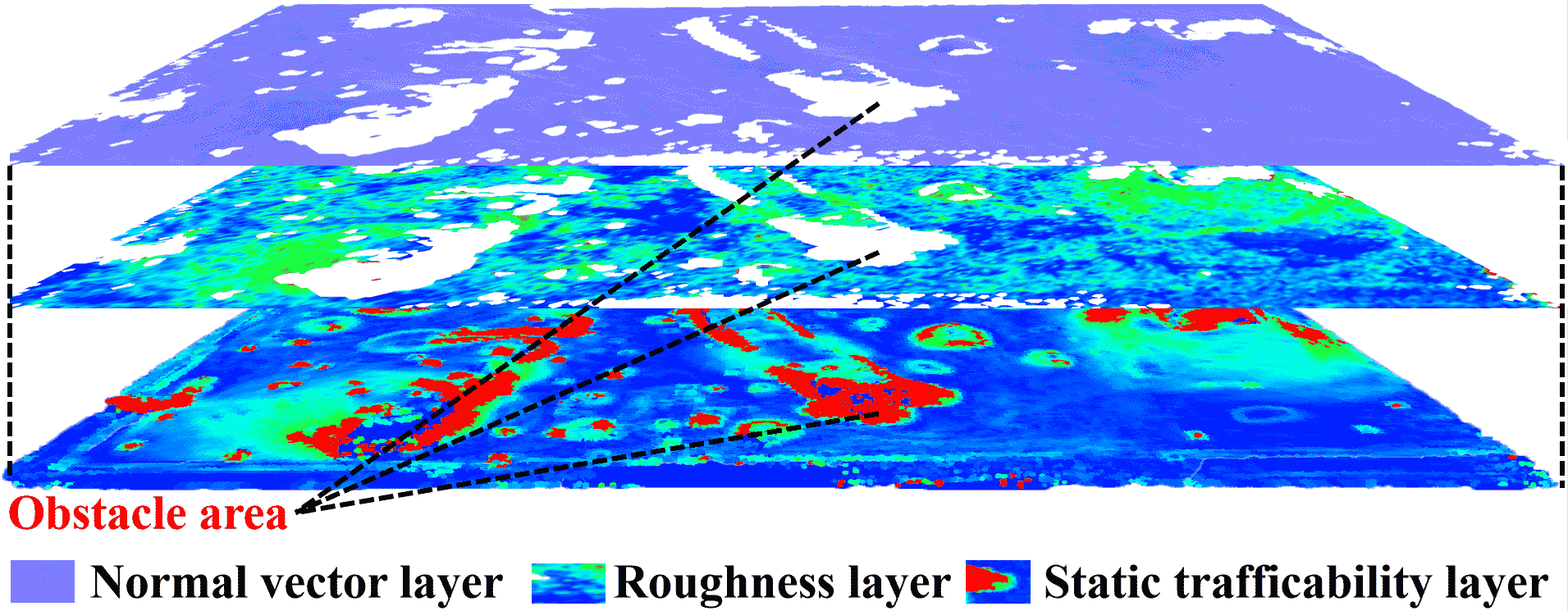}
  \caption{Hybrid map with multiple layers of information. From top to bottom of the picture are the normal vector layer, the roughness layer, and the static traversability layer, respectively.
  The red blocks of static traversability layer represent the obstacles and are stored as the 2D map. The white blocks of normal vector layer and roughness layer indicate areas where there is no normal vector or roughness information, meaning that the 2D map only has single-layer static traversability information.}
  \label{fig:layer}
\end{figure}

The maximum height difference $h$ of point in the small voxel is calculated by
\begin{equation}
h = \max_{i=1,2,...n}(d_i)-\min_{j=1,2,...n}(d_j) \label{5},
\end{equation}
where $d_i$ is the distance from $\v{p_i}$ to the fitted plane, and $n$ is the number of points in the small voxel grid. 
The distances are signed so that pits and bumps can be detected. 
Considering the obstacle-crossing ability of the robot, the maximum height difference threshold is set to $h_{max}$. 
If $h > h_{max}$, the area of the small voxel grid is stored as a 2D obstacle map containing only one layer of information with zero static traversability. 
Otherwise, it is stored as a 2.5D DEM with multi-layer terrain information. 
Then, as shown in~\figref{fig:layer}, the hybrid map is generated. 
The obstacle areas belong to the 2D map with only one layer of static traversability information, and the areas are marked in red with zero traversability. 
The other areas belong to 2.5D DEM, and our DEM includes the normal vector of the fitted plane, the terrain roughness $r_{sum}$, and static terrain traversability $\tau \in [0,1]$. $r_{sum}$ is the sum of the absolute values of residuals in the small voxel grid as
\begin{equation}
r_{sum} = \sum^{n}_{k=1}|d_k| \label{6}.
\end{equation}
In the area where static traversability $\tau$ is zero, the robot cannot pass through in any orientation.  
Otherwise, $\tau$ is calculated based on roughness and slope as 
\begin{equation}
\tau = 1-W_r \frac{r_{sum}}{r_{max}}-W_{\rho} \frac{\rho}{\rho_{max}} \label{7},
\end{equation}
where $W_r$ and $W_{\rho}$ are both weighting coefficients. 

In the path planning process, the static traversability layer is used to quickly remove impassable nodes. Then the real terrain traversability will be calculated according to the terrain roughness and the robot's 3D orientation projected on the ground.

\subsection{T-Hybrid A*}
\label{sec:tha}
After generating the hybrid map with traversability information of the unstructured environment, we then modify the Hybrid A* algorithm by using traversability as an extra safety constraint.

Hybrid A* algorithm~\cite{petereit2012robotik} considers kinematic constraints of the robot, and thus the planned path is easier for the robot to track than A* algorithm~\cite{hart1968ssc}. 
However, for robots in unstructured outdoor environments with slopes and other rough terrains, the path planned by Hybrid A* may pass through low-traversability areas.
In this case, the robot may face difficulties or even the risk of capsizing. 
To deal with this problem, we propose T-Hybrid A*, where the terrain traversability is induced as a new safety constraint.

For a given hybrid map, start pose, and goal pose in the $SE(2)$, a safe path is generated by T-Hybrid A*. 
In node expansion, if the child node is located on the 2D map, it would be directly deleted since its corresponding static traversability is zero. 
Otherwise, the child node will be queried on the 2.5D map, wherein the robot orientation on the terrain is calculated according to its heading angle.
Assuming the node is $[x, y,\theta] \in SE(2)$, the fitted plane normal vector $\v{v}$ can be queried according to the node position. The node heading angle is mapped to the orientation $\m{R}_{MR}$ in global map frame $M$ by
\begin{equation}
\begin{small}
\m{R}_{MR} = [\widehat{\v{x}}_{MR} \quad \widehat{\v{y}}_{MR} \quad \widehat{\v{z}}_{MR}] = {
\left[ \begin{array}{ccc}
\cos \theta & \sin \theta & 0\\
-\sin \theta & \cos \theta & 0\\
0 & 0 & 1
\end{array} 
\right ]} \label{8},
\end{small}
\end{equation}
where frame $R$ is parallel to the gravity coordinate frame, and $[\widehat{\v{x}}_{MR} \quad \widehat{\v{y}}_{MR} \quad \widehat{\v{z}}_{MR}]$ is the vector representation of orientation. 
$\v{v}$ is the z-axis of projection frame $\widetilde{R}$.
Due to the condition of zero yaw between $\m{R}_{MR}$ and the robot's projection orientation $\m{R}_{M\widetilde{R}}$, the x-axis of frame $\widetilde{R}$ is perpendicular to the y-axis of frame $R$. 
Therefore, $\m{R}_{G \widetilde{R}}$ can be calculated by the following formula:  
\begin{equation}
\widehat{\v{x}}_{M\widetilde{R}} = \frac{\widehat{\v{y}}_{MR} \times \v{v}}{\|\widehat{\v{y}}_{MR}  \times \v{v} \| } \label{9}
\end{equation}
\begin{equation}
\m{R}_{M\widetilde{R}} = [\widehat{\v{x}}_{M\widetilde{R}} \quad \v{v} \times \widehat{\v{x}}_{M\widetilde{R}} \quad \v{v}]   \label{10}
\end{equation}

We then transform the rotation matrix to Euler angles in Z-Y-X rotation order to obtain the robot's roll angle and pitch angle $[\theta_x, \theta_y,\theta_z]$.
The real terrain traversability $\widetilde{\tau}$ is calculated from the terrain roughness, the real pitch and the roll angle of the robot as
\begin{equation}
\widetilde{\tau} = 1-w_r \frac{r_{sum}}{r_{max}}-w_{\theta x} \frac{|\theta_x|}{\theta_{xmax}}-w_{\theta y} \max(\frac{\theta_y}{\theta_{ymin}},\frac{\theta_y}{\theta_{ymax}})   \label{11},
\end{equation}
where each item is normalized. 
$r_{max}$ and $\theta_{xmax}$ are max roughness and roll angle safety threshold, respectively. 
Considering the asymmetry between the front and the back of the robot, the safety value of the robot going uphill and downhill is different, so there are two different limit values ($\theta_{ymin}$ and $\theta_{ymax}$) for the pitch. 
During path planning, path nodes expand forward according to Ackerman kinematic model, and the node cost $f(s_{i+1})$ is calculated by
\begin{equation}
    f(s_{i+1}) = 
   \begin{cases}
     \infty & \hspace{-15mm} \mbox{$d_o$ = 0 or $\widetilde{\tau}$ = 0}\\
     f(s_i) + k_t t_{i+1}+k_{\widetilde{\tau}}\widetilde{\tau} & \mbox{others}
   \end{cases},
\end{equation}
where $d_o$ is the distance to the nearest obstacle in configuration space. 
If $d_o$ or terrain traversability $\widetilde{\tau}$ is 0, the cost of the node will be infinite. 
Otherwise, the cost of the current node is the sum of the cost of the parent node $f(s_i)$, turning cost $t_{i+1}$, and traversability cost $\widetilde{\tau}$. 
$k_t$ and $k_{\widetilde{\tau}}$ are proportional factors. 
Through the graph search method, a safe path can be efficiently found in the unstructured outdoor environment.


\section{Experimental Evaluation}
\label{sec:exp}

\begin{figure}[t]
  \centering
  \includegraphics[width=0.75\linewidth]{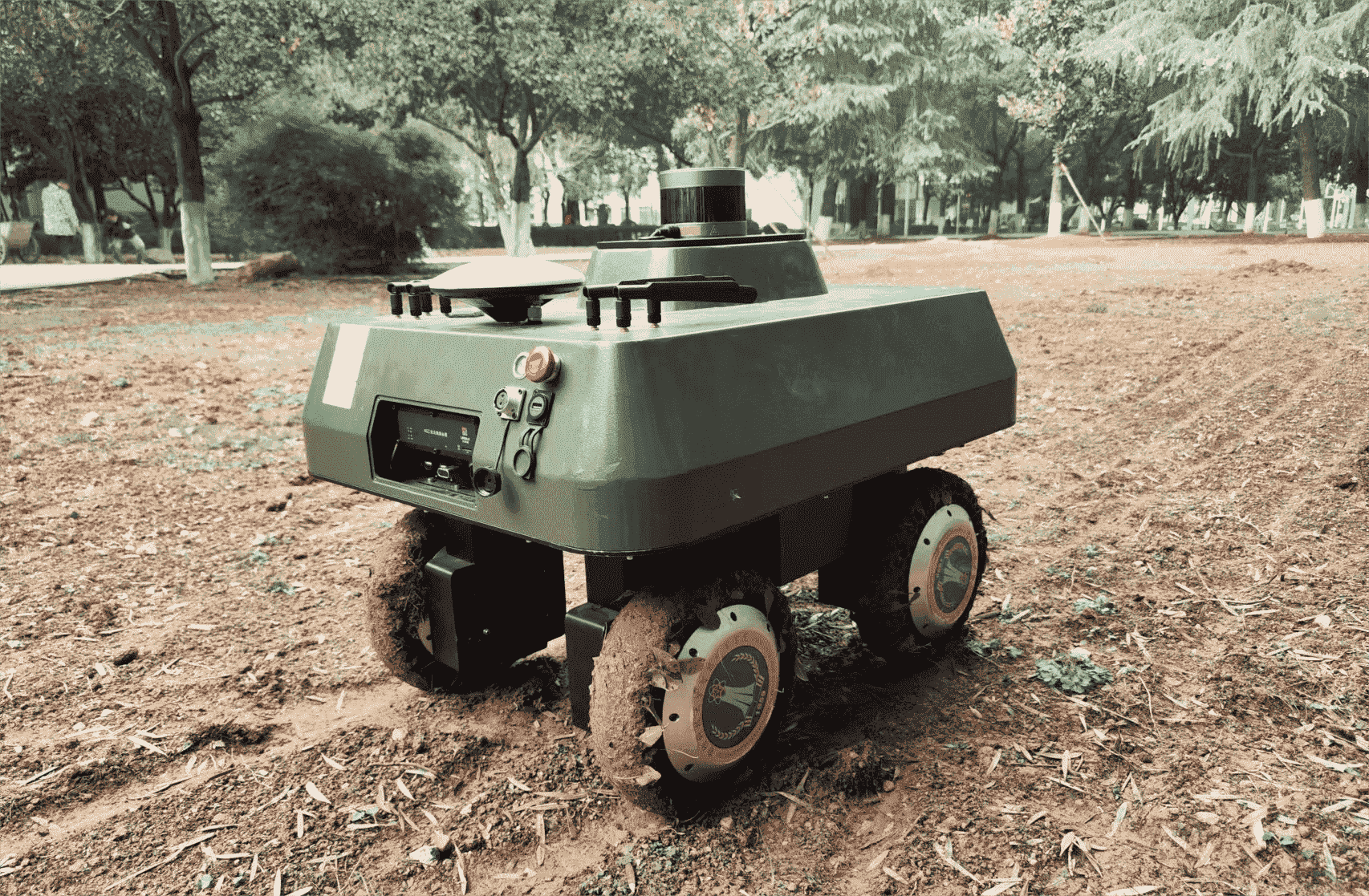}
  \caption{Our real robot platform works in unstructured terrain. The robot is equipped with an i5 processor, a Velodyne 16 Lidar sensor, and a fusion positioning system.}
  \label{fig:robot}
\end{figure}

The main focus of this work is a path-planning method for the ground robot in unstructured outdoor environments. Our method builds a 2D and 2.5D fused map and leverages a novel path planning to generate a short and safe path in less time.

Experiments have been conducted in both simulator and real outdoor unstructured terrains to evaluate the proposed method, and the results can support our key claims, which are:
(i) Our method can assess various complex terrain accurately, and create a hybrid map representation possessing the advantages of 2D and 2.5D maps.
(ii) Our proposed path planning method based on our hybrid map generates safer paths than that based on the 2D map. Moreover, our method takes less time and generates shorter paths than the 2.5D-based method;
(iii) Our path planning method can be used on the physical robot to navigate in unstructured outdoor environments.

\subsection{Implementation Details and Experimental Setup}   

\begin{table}[t]
  \caption{Parameters in the Experiments}
  \centering
  \renewcommand\arraystretch{1.2}
  \setlength{\tabcolsep}{6pt}
  \begin{tabular}{cc|cc}
    \toprule
      parameter & value & parameter   & value  \\
    \midrule
     $R_l$ &  0.7 $m$ & $\theta_{xmax}$  &  0.18 $rad$\\
     $R_w$ &  0.5 $m$ & $\theta_{ymin}$  &  -0.3 $rad$\\
      $\rho_{max}$  &  0.3 $rad$ & $\theta_{ymax}$  &  0.25 $rad$\\
      $h_{max}$  &  0.1 $m$ & $w_{\theta x}$  & 0.4\\
      $W_r$  &  0.5 & $w_{\theta y}$  & 0.4\\
      $W_{\rho}$ &  0.5  & $k_t$ & 1.1\\
      $w_r$ & 0.3 & $k_{\widetilde{\tau}}$ & 2.0\\
    \bottomrule
    \label{tab:1}
  \end{tabular}

  \label{tab:parameter}
\end{table}

\begin{figure}[t] 
	\centering  
	\subfigbottomskip=2pt 
	\subfigcapskip=-5pt 
	\hspace{-3mm}
	\subfigure[]{
		\label{data.sub.1}
		\includegraphics[width=0.49\linewidth]{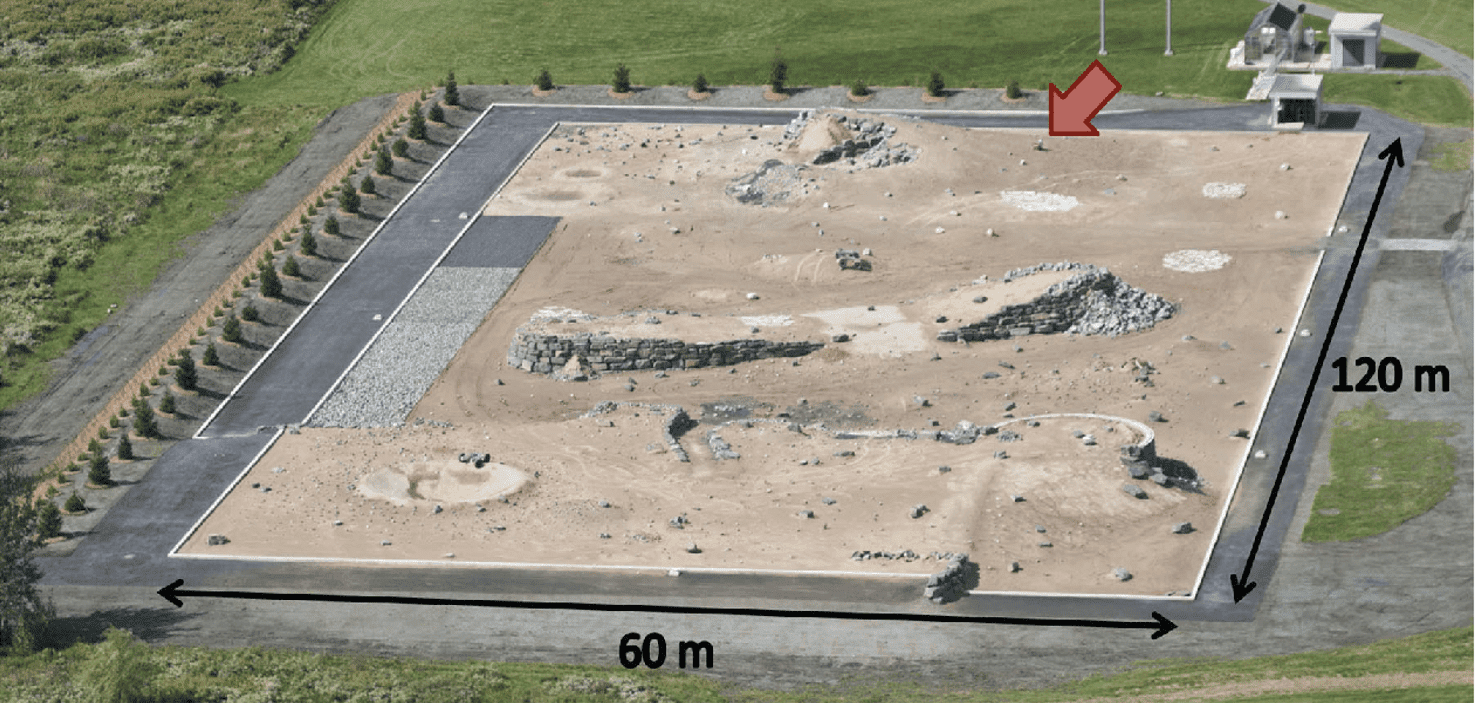}}
     \hspace{-3mm}  
	\subfigure[]{
		\label{data.sub.2}
		\includegraphics[width=0.49\linewidth]{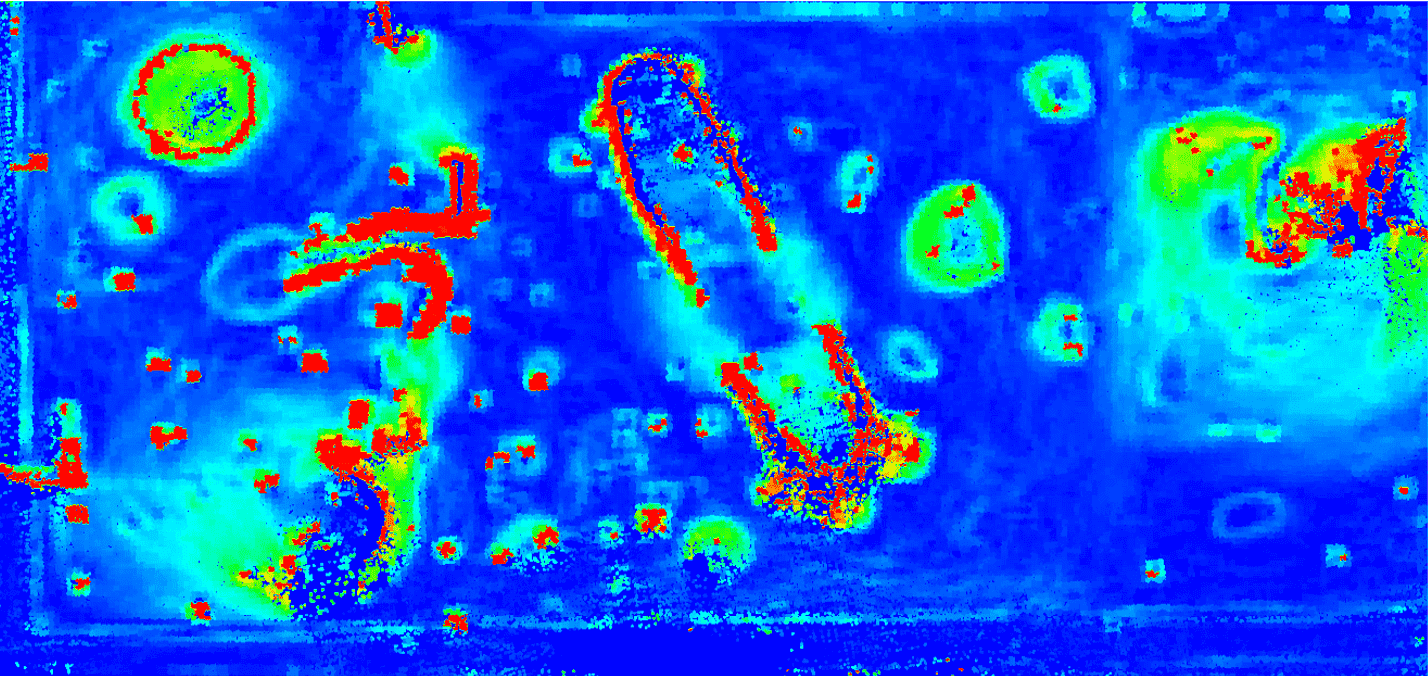}}
        \\
    \hspace{-3mm}
	\subfigure[]{
		\label{data.sub.3}
		\includegraphics[width=0.49\linewidth]{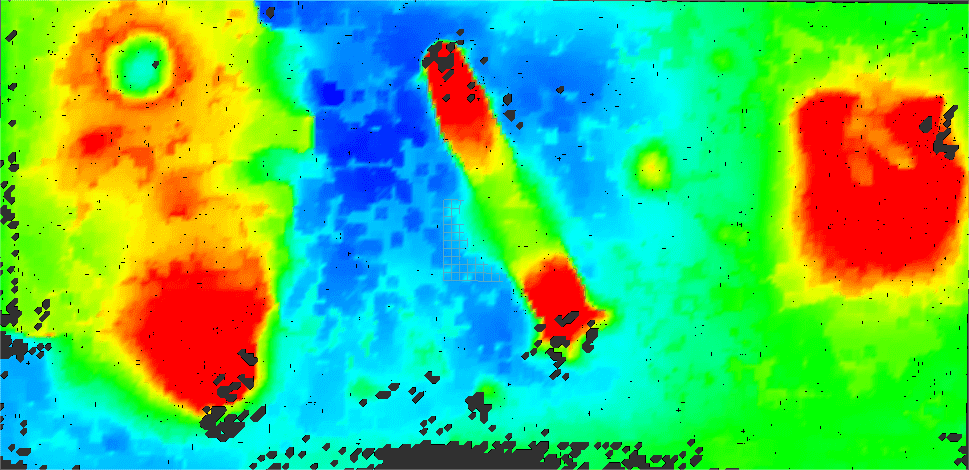}}
     \hspace{-3mm}  
	\subfigure[]{
		\label{data.sub.4}
		\includegraphics[width=0.48\linewidth]{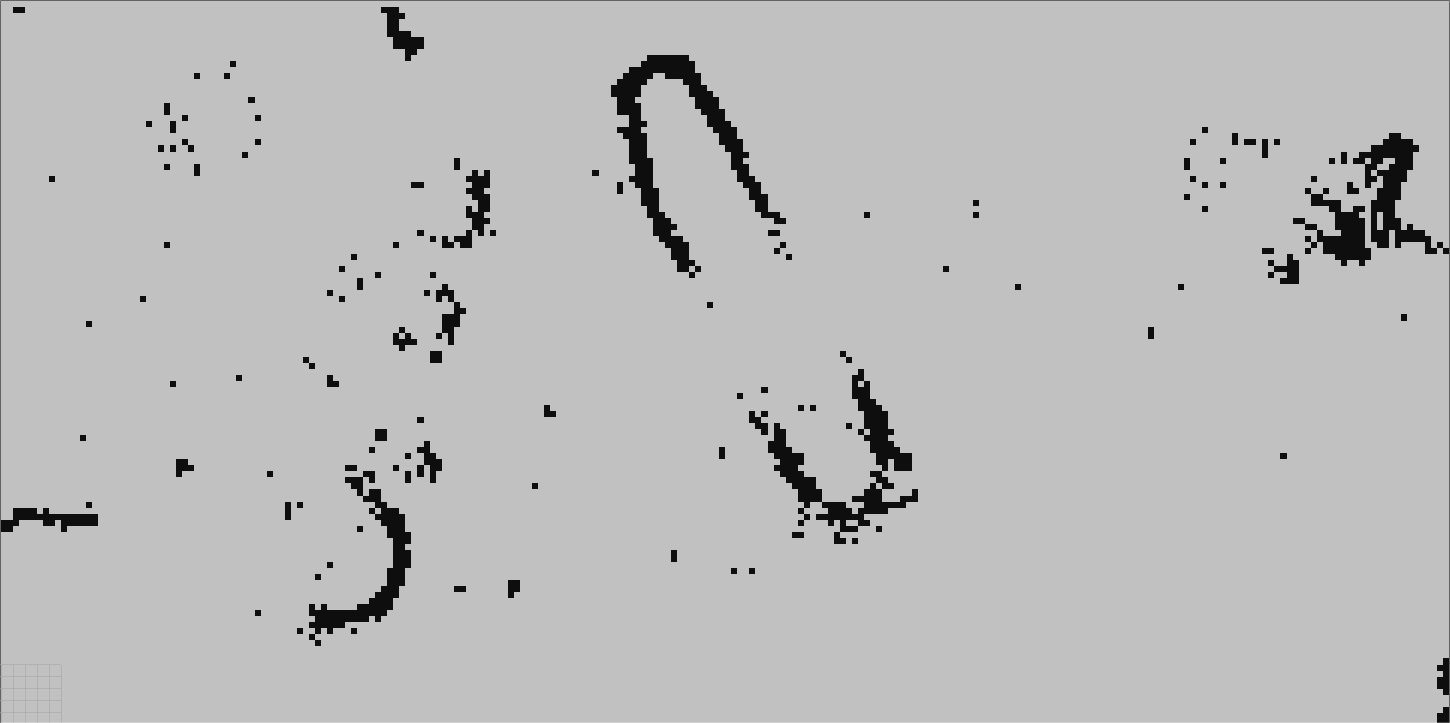}}
	\caption{Terrain assessment of Mars-like terrain data set based on three maps. (a) is the aerial view of Mars-like terrain, and the terrain size is $120m \times 60m$. (b) is the hybrid map generated by our terrain assessment algorithm, and its color represents static traversability. The traversability gradually increases when the color ranges from warm to cool. The red color represents the impassable area, which is stored as the 2D map, and the rest areas are stored as the 2.5D map. (c) and (d) are 2.5D elevation map and 2D occupancy grid map respectively.}
	\label{data}
 \vspace{-0.2cm}
\end{figure}

\begin{figure*}[htb]
	\centering  
	\subfigure[]{
		\label{assess.sub.1}
		\includegraphics[width=0.335\linewidth]{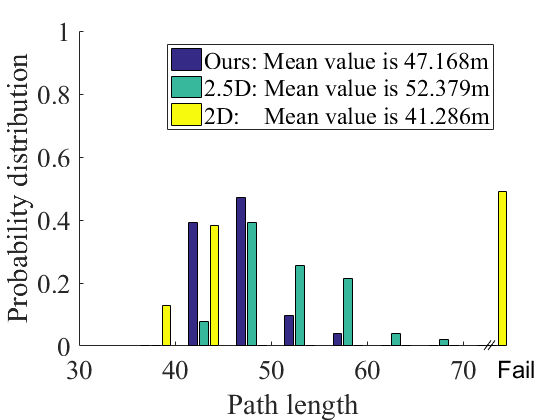}}
     \hspace{-5mm}  
	\subfigure[]{
		\label{assess.sub.2}
		\includegraphics[width=0.335\linewidth]{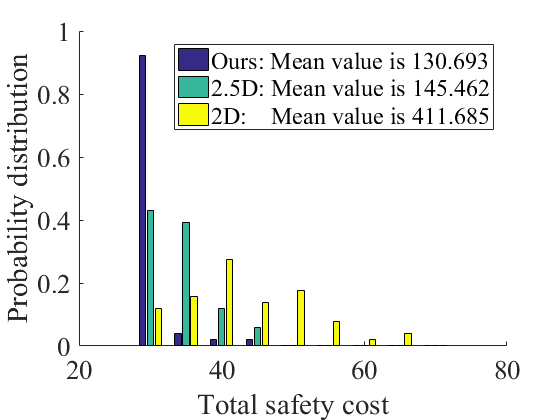}}
    \hspace{-5mm}  
	\subfigure[]{
		\label{assess.sub.3}
		\includegraphics[width=0.335\linewidth]{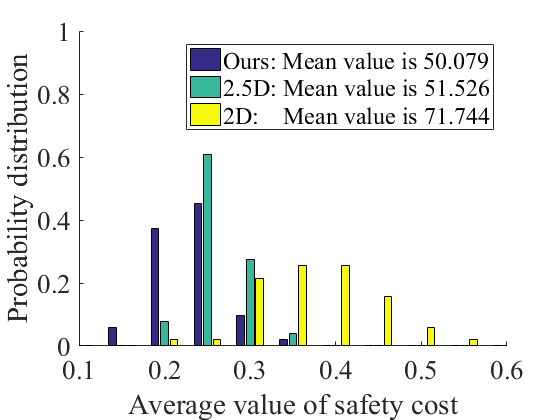}}
	\caption{Comparative analysis of 50 random experiments on the dataset, and the ordinate of each picture represents the distribution probability. (a) represents the comparison of three methods in the planned path length. At the same time, the mean value of each method is indicated in the upper right corner of the image. Similarly, (b) is the total safety cost distribution. In order to eliminate the influence of path length on the total safety cost, we calculate the average value of the safety as shown in (c).}
	\label{assess}
 \vspace{-0.2cm}
\end{figure*}

The raw 3D point cloud map is constructed  using a SLAM system \cite{shan2020iros}, and the elevation map is generated  following Fankhauser’s method \etalcite{fankhauser2016universal}. The elevation map is then transformed into a 2.5D digital elevation map with a fitted plane normal vector, roughness, and static traversability. 
Our terrain assessment method subsequently generates a hybrid map, Finally, the three kinds of map information are combined with T-Hybrid A* for path planning. 
The map resolution and threshold are unified according to the robot performance, and all parameters are shown in~\tabref{tab:1}.

In the following sections, we will compare our method against two different baseline methods: a 2D map-based method \cite{dolgov2008practical} and a 2.5D map-based method \cite{Wermelinger2016iros}, using both the Canadian Mars Simulation surface dataset~\cite{tong2013ijrr} and the physical robot working in real outdoor unstructured terrains.
The evaluation metrics include the path length, safety cost, and the planning time. 
As shown in~\figref{fig:robot}, our outdoor robot is with the Ackerman-like kinematic mode, and its size is $0.7m \times 0.5m \times 0.5m$. 
The robot is equipped with an i5 processor, a Velodyne 16 Lidar sensor, and a GNSS-IMU fusion positioning system. 
The robot has a certain ability to cross $0.1m$ bumps, $-0.1m$ pits, and $20^{\circ}$ slopes.

\subsection{Qualitative Experiment on Publicly Available Dataset}

\begin{figure}[t]
  \centering
  \includegraphics[width=1.0 \columnwidth]{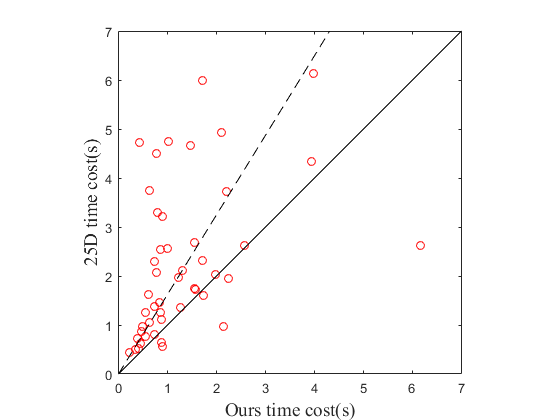}
  \caption{The time cost distribution of 50 path planning experiments on the Mars-like terrain dataset. Each red dot represents a pair of starting and ending points, the abscissa of this dot represents the time cost of our path planning method under this pair of points, and the ordinate is the time cost of planning based on 2.5D map.}
  \label{fig:tims}
  \vspace{-0.2cm}
\end{figure}

We first show the experimental results on the publicly available Canadian Planetary Emulation Terrain 3D mapping dataset~\cite{tong2013ijrr}. 
\figref{data.sub.1} shows the overview of the simulated scenario. It is a $120m \times 60m$ unstructured outdoor environment with pits, stone bumps, slopes, and other complex terrains. 
The hybrid map, 2.5D digital elevation map, and 2D occupancy grid map are generated by our terrain assessment method, Fankhauser's terrain assessment methods, respectively.
\figref{data.sub.2} - \figref{data.sub.4} show different maps with a uniform resolution of $0.5m$. The color of the map in \figref{data.sub.2} represents the static terrain traversability $\tau$, the red is the absolutely impassable 2D map area, while the rest areas are 2.5D areas, and with the color changes from warm to cool, the static terrain traversability gradually increases.

We randomly sample positions in the map with a 40m distance between the start and goal points, and discard any impassable start or goal points.
Our T-Hybrid A* carries out 50 path planning experiments on three kinds of maps, and we analyzes the path length, safety cost, and planning time, respectively. 
The safety cost is measured by the hazard value, which is a synthesis of the robot's pitch angle, roll angle, and terrain roughness obtained from all path points. 
A higher hazard value indicates lower safety.
Note that path planning based on the 2D map does not consider the robot's pose, which sacrifices the safety in exchange for a short planning time. 
Therefore, in terms of planning time, we only compared the method based on the 2.5D map with our method.

The evaluation results of three different setups are shown in~\figref{assess}. \figref{assess.sub.1} illustrates the distribution of path lengths planned by three methods, and the ordinates are their respective distribution probabilities. 
As can be seen, half of the paths planned based on 2D maps are failed and cannot be safely tracked, because the terrain roughness or robot projection orientation at some of the path points exceeds the corresponding safety thresholds.
Meanwhile, the paths generated by our method and the 2.5D-based method can be safely tracked, while ours are shorter than those of 2.5D method. 
The safety analysis of generated paths in \figref{assess.sub.2} shows that our paths are the safest, while the paths generated by the 2D-based method are the most impassable. 
Furthermore, in addition to the total safety cost, we also provide the corresponding mean value in \figref{assess.sub.3}. 
Compared to methods based on the 2D map and 2.5D map-based methods, our planning method has better results in path length and safety.
To further compare our method with the 2.5D-based method, we provide a time cost analysis in \figref{fig:tims}, where each red dot represents a pair of fixed starting point and goal point. 
The abscissa of the dot represents the time cost of our path planning method under this pair of points, and the ordinate is the time cost of planning based 2.5D map. 
The dashed line clearly illustrates that our method exhibits a statistically significant improvement over the 2.5D-based path-planning approach in terms of time cost.






\subsection{Ground Vehicle Physical Experiment}

\begin{figure*}[t]
	\centering  
    \hspace{-5mm}  
	\subfigure[]{
		\label{terrain1.sub.1}
		\includegraphics[width=0.48\linewidth]{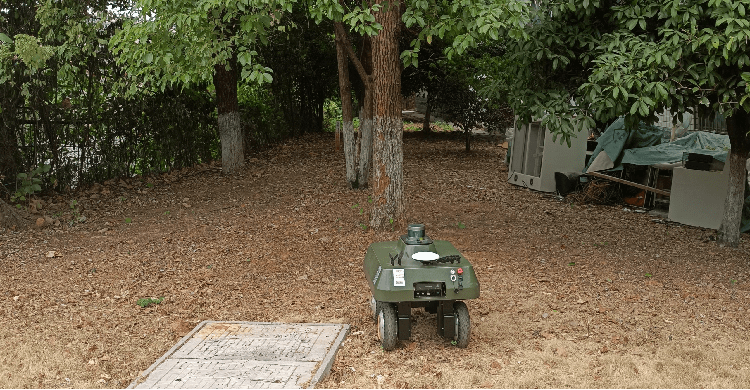}}
     \hspace{-3mm}  
	\subfigure[]{
		\label{terrain2.sub.1}
		\includegraphics[width=0.48\linewidth]{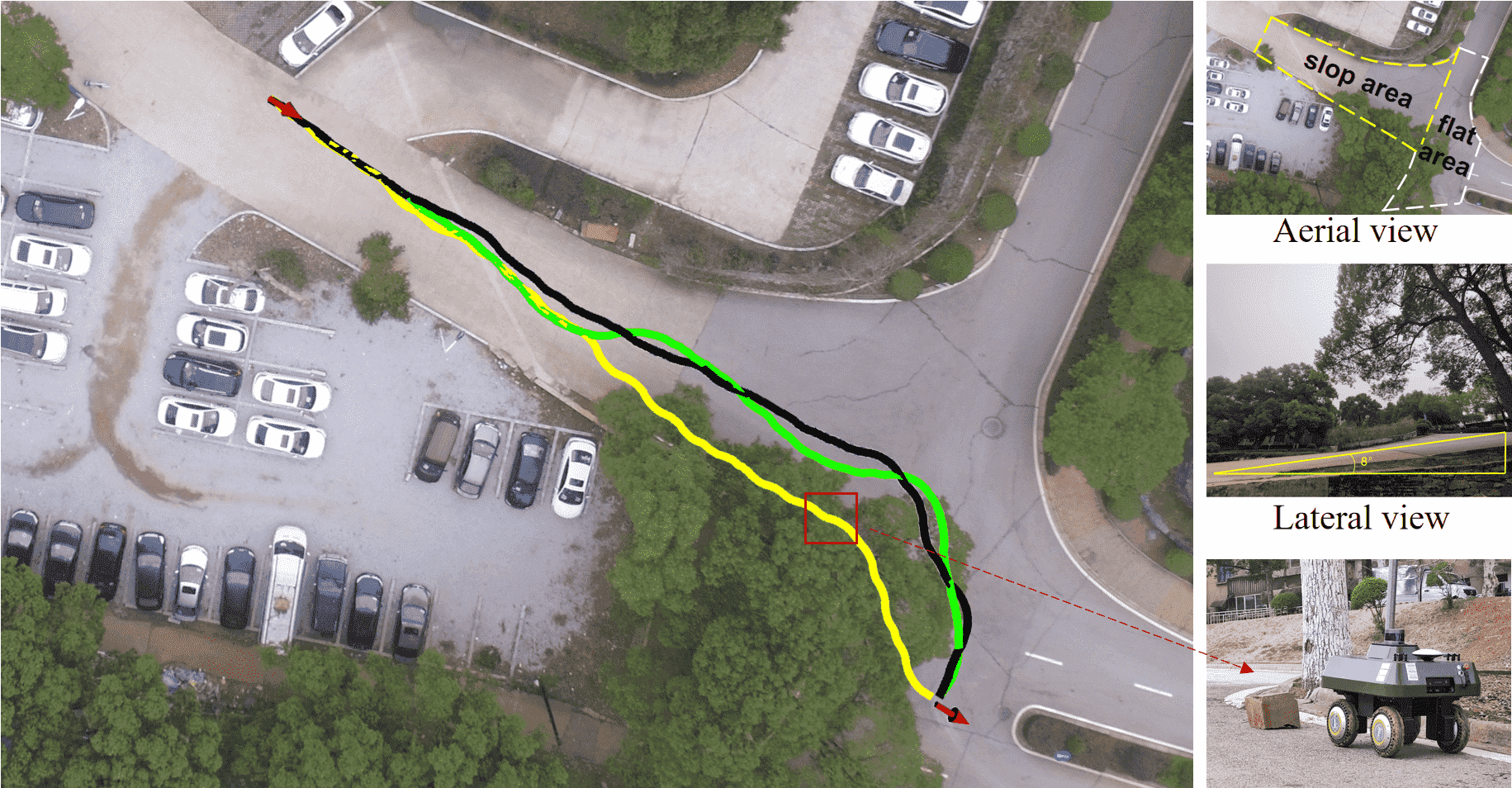}}
       \\
       \hspace{-5mm} 
	\subfigure[]{
		\label{terrain1.sub.2}
		\includegraphics[width=0.48\linewidth]{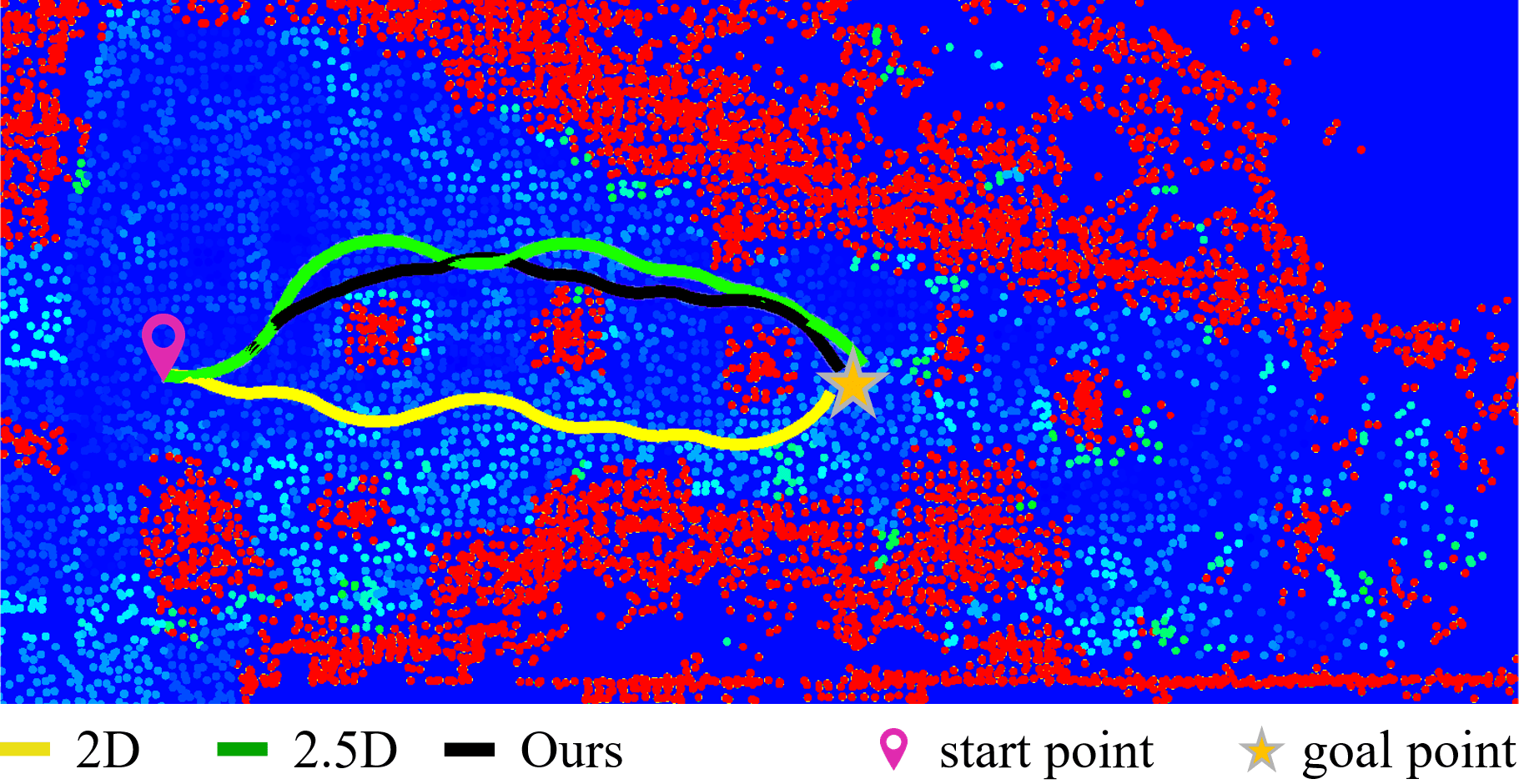}}
     \hspace{-3mm}  
	\subfigure[]{
		\label{terrain2.sub.2}
		\includegraphics[width=0.48\linewidth]{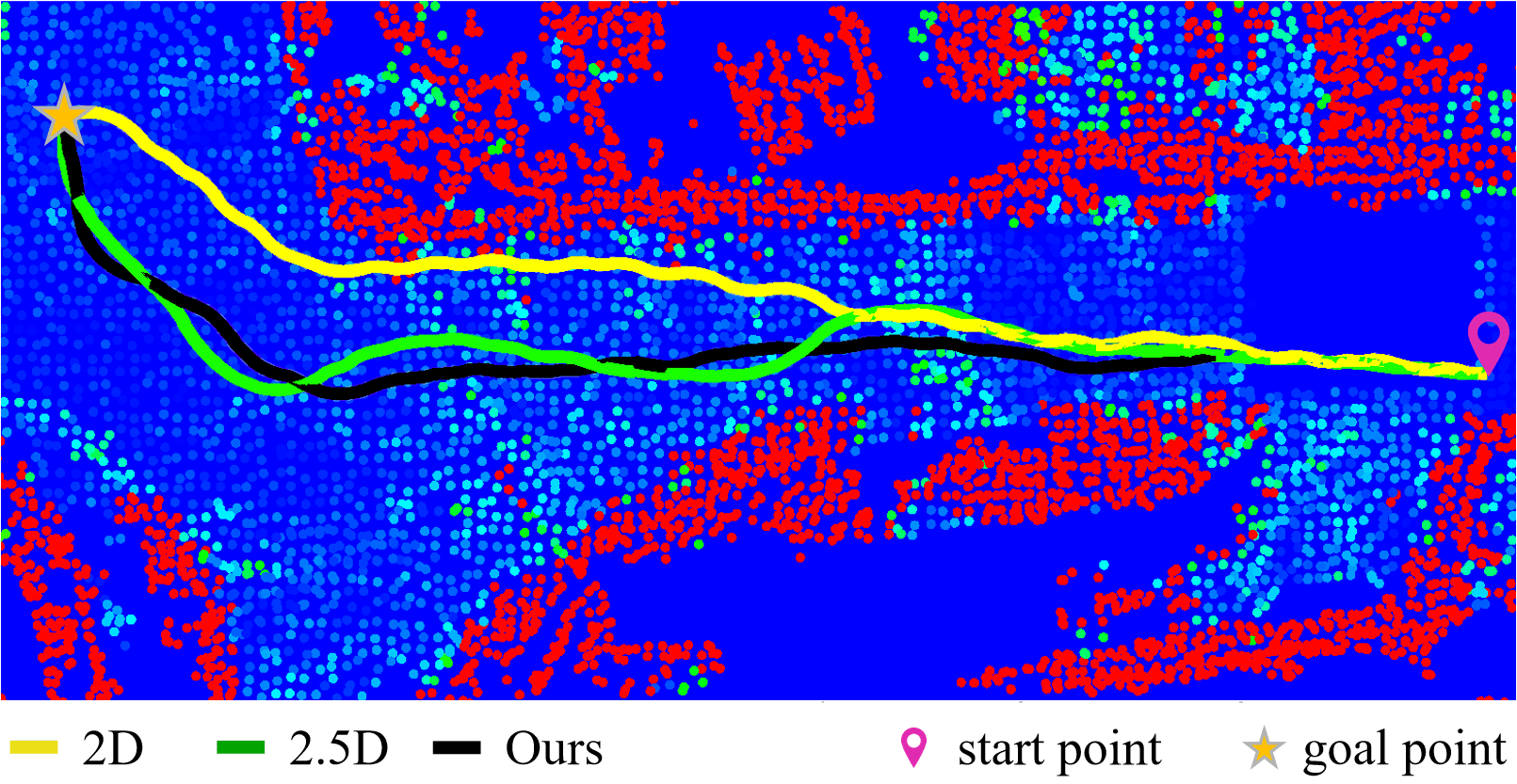}}
	\caption{Path planning with three different methods in unstructured outdoor terrains. (a) shows the actual rough terrain. On the right side of the image is a small slope. (b) shows a long slope, and there is a big stone at the right top of the slope. (c) and (d) are hybrid maps of the above terrains. It can be seen from the figure that the static traversability of the slope area is lower in green, while the flat area is blue. All three paths are drawn in our hybrid map, and the black path is the resulting planning based on our hybrid map, while the green and yellow paths are generated based on the 2.5D map and 2D map, respectively.}
	\label{terrain}
\end{figure*}

\begin{table}[t]
  \caption{Analysis of navigation results in outdoor rough areas}
  \centering
  \renewcommand\arraystretch{1.2}
  \setlength{\tabcolsep}{2.4pt}
  \begin{tabular}{C{0.7cm}|c|cccccc}
    \toprule
      & Methods & Time(s) & Length(m)   & Roll\,cost   & Pitch\,cost  &  Failure\,rate \\
    \midrule
     \multirow{3}{*}{\rotatebox[origin=c]{90}{Terrain 1}}& 2D  &  - & - & 411.685  & 71.744  & 0.319\\                
     & 2.5D &  3.359 &  13.139  & 145.462  & 51.526 & 0.000 \\
     & Ours  &  2.738 &  12.290  & 130.693  & 50.079 & 0.000 \\
     \midrule
     \multirow{3}{*}{\rotatebox[origin=c]{90}{Terrain 2}}& 2D  &  - & - & -  & -  & -\\ 
     & 2.5D &  0.954 &  51.081  & 537.349  & 337.115 & 0.000 \\
     & Ours  &  0.675 &  50.255  & 464.061  & 325.798 & 0.000 \\
    \bottomrule
  \end{tabular}
  \vspace{-0.2cm}
  \label{Physical_experiment}
\end{table}

When the robot moves in unstructured outdoor environments, in addition to avoiding obstacles, it should try to avoid rough and slope areas. At the same time, if the robot has to pass a long slope, it will be safer for the robot to pass in forward orientations.
In order to verify the actual effect of our method, we conducted experiments on a real physical robot. We conducted navigation experiments in a rough and un-flat terrain 1 (see~\figref{terrain1.sub.1}) and a long slope terrain 2  (see~\figref{terrain2.sub.1}). \figref{terrain1.sub.2} and \figref{terrain2.sub.2} show the path planning results visualized on the hybrid maps generated using our terrain traversability assessment.
As can be seen, the static traversability of the long slope area is lower in green color, while the flat area is blue. We set a unified start pose and target pose. The yellow, green, and black paths are generated based on 2D, 2.5D, and our hybrid map.
As shown in~\figref{terrain1.sub.2}, our path is the safest for avoiding the sloping area on the right of terrain 1. Besides, our path is shorter than 2.5D's. 
When the robot has to cross a long slope as shown in~\figref{terrain2.sub.1}, since our method considers the robot's 3D orientation, our path crosses the slope in the forward direction. On the contrary, the 2D-based method failed because its path crosses the red obstacle area.

The same MPC algorithm is exploited for the robot to track the above different paths. The quantitative comparison results are shown in~\tabref{Physical_experiment}.
Roll cost and Pitch cost represent the sum of the normalized roll angle and normalized pitch angle in the path-tracking process. 
Our path consists of many path points. 
To ensure the robot's safety, we treat waypoints exceeding the corresponding parameter's threshold as failure points.
These parameters include terrain roughness, pitch angle, and roll angle of the robot on this terrain. 
The failure rate is the ratio of failure points at all path points.


Our method outperforms other methods in terms of both efficiency and safety for the above-valuated terrains. Specifically, in Terrain 1, the 2D-based path planning method passes through the slope area resulting in the highest robot orientation changes, while our method exhibits the lowest roll angle changes, indicating a significant advantage over other baseline methods. In Terrain 2, the 2D-based method fails to plan a feasible path as it crosses an obstacle and passes through a long slope in a poor orientation. In contrast, our method results in the minimum pitch and roll angle changes during the path tracking process. Terrain 1 is generally rougher and more complex than Terrain 2, thus requiring more iterations in path planning. Our method achieves lower planning time consumption and higher safety scores than the 2.5D-based method in both terrains.

\section{Conclusion}
\label{sec:conclusion}

In this paper, we presented a novel approach to efficiently plan a safe path for the ground robot in unstructured outdoor environments.
To deal with overhanging structures, dynamic voxel grids have been employed to segment the raw point clouds.
Afterwards, a hybrid map is generated combining the advantages of 2D occupancy grid map and 2.5D digital elevation map.
Based on the hybrid map, we proposed a novel path planning algorithm called T-Hybrid A*, which considers the robot's kinematics, obstacle distance, and terrain traversability and efficiently together. 
As a result, the robot can navigate in unstructured outdoor environments with slopes, bumps, and pits. 
Our approach has been evaluated on a simulated dataset and a real robot platform in unstructured environments. 
We also provided comparisons against other existing techniques and supported all claims made in this paper. 
The experiments suggest that the path planned by our method is safer than the path planned based on 2D map, shorter and less time-consuming than 2.5D's. 
In the future, we will focus on path planning at larger outdoor unstructured environments.



\bibliographystyle{IEEEtran}
\bibliography{paper}


\end{document}

%% file: stachnisslab-latex.tex
\usepackage{graphics}           
\usepackage{times}              
\usepackage{amsmath}            
\usepackage{amssymb}            
\usepackage{graphicx}
\usepackage{algorithm}
\usepackage[noend]{algpseudocode}
\usepackage{booktabs}
\usepackage{color}
\definecolor{instructioncolor}{rgb}{.5,.5,.5}

\usepackage[font=small]{caption}

\def\secref#1{Sec.~\ref{#1}}
\def\figref#1{Fig.~\ref{#1}}
\def\tabref#1{Tab.~\ref{#1}}
\def\eqref#1{Eq.~(\ref{#1})}

\makeatletter
\usepackage{xspace}
\DeclareRobustCommand\onedot{\futurelet\@let@token\@onedot}
\def\@onedot{\ifx\@let@token.\else.\null\fi\xspace}

\def\etal{{et al}\onedot}
\makeatother

\def\etalcite#1{\etal~\cite{#1}}

\usepackage{array}
\newcolumntype{L}[1]{>{\raggedright\let\newline\\\arraybackslash\hspace{0pt}}m{#1}}
\newcolumntype{C}[1]{>{\centering\let\newline\\\arraybackslash\hspace{0pt}}m{#1}}
\newcolumntype{R}[1]{>{\raggedleft\let\newline\\\arraybackslash\hspace{0pt}}m{#1}}

%% file: stachnisslab-math.tex
\renewcommand{\b}[1]{\mbox{\boldmath$#1$}}

\renewcommand{\v}[1]{{\b #1}} 

\newcommand{\m}[1]{{\mbox{{\sffamily\slshape{#1\/}}}}}

